\documentclass[letterpaper, 10 pt, conference]{ieeeconf}  

\IEEEoverridecommandlockouts                              

\usepackage{graphics} 
\usepackage{epsfig} 
\usepackage{mathptmx} 
\usepackage{times} 
\usepackage{amsmath} 
\usepackage{amssymb}  

\usepackage{xspace}  
\usepackage{tabularx}  
\usepackage{booktabs}  
\usepackage{array}  
\usepackage{xcolor}  
\usepackage{subcaption}  
\usepackage{multirow}  
\usepackage{makecell}  
\usepackage{pgfplots}  
\usepackage{acro}  
\usepackage{placeins}  
\usepackage[colorlinks=true]{hyperref}  
\usepackage{fancyhdr}

\pgfplotsset{compat=newest}
\graphicspath{{./graphics/}}

\newcommand{\figref}[1]{Fig.~\ref{#1}}
\newcommand{\secref}[1]{Section~\ref{#1}}

\newcommand{\tabref}[1]{Table~\ref{#1}}

\makeatletter
\DeclareRobustCommand\onedot{\futurelet\@let@token\@onedot}
\def\@onedot{\ifx\@let@token.\else.\null\fi\xspace}

\def\eg{e.g\onedot}

\def\etal{et~al\onedot}

\makeatother

\newcolumntype{x}[1]{>{\centering\arraybackslash\hspace{0pt}}p{#1}}

\DeclareAcronym{ap}{short=ap, long=average precision}
\DeclareAcronym{bev}{short=BEV, long=bird's-eye view}
\DeclareAcronym{CD}{short=CD, long=\textit{Chamfer} distance}
\DeclareAcronym{CNN}{short=CNN, long=convolutional neural network}
\DeclareAcronym{DA}{short=DA, long=domain adaptation}
\DeclareAcronym{DNN}{short=DNN, long=deep neural network}
\DeclareAcronym{EMD}{short=EMD, long=\textit{Earth Mover's} distance}
\DeclareAcronym{FCN}{short=FCN, long=fully convolutional network}
\DeclareAcronym{FID}{short=FID, long=Fr\'{e}chet Inception Distance}
\DeclareAcronym{FOV}{short=FOV, long=field of view}
\DeclareAcronym{FPD}{short=FPD, long=Fr\'{e}chet point cloud distance}
\DeclareAcronym{GAN}{short=GAN, long=generative adversarial network}
\DeclareAcronym{GCA}{short=GCA, long=Geodesic Correlation Alignment}
\DeclareAcronym{iou}{short=IoU, long=intersection over union}
\DeclareAcronym{IS}{short=IS, long=Inception Score}
\DeclareAcronym{KL}{short=KL, long=\textit{Kullback-Leibler}}
\DeclareAcronym{lidar}{short=LiDAR, long=light detection and ranging}
\DeclareAcronym{MAE}{short=MAE, long=mean absolute error}
\DeclareAcronym{MLP}{short=MLP, long=multi-layer perceptron}
\DeclareAcronym{MOS}{short=MOS, long=mean opinion score}
\DeclareAcronym{MSE}{short=MSE, long=mean squared error}

\newcommand{\kitti}{\textit{KITTI}}
\newcommand{\semantickitti}{\textit{SemanticKITTI}}
\newcommand{\nuscenes}{\textit{nuScenes}}
\newcommand{\semanticposs}{\textit{SemanticPOSS}}
\newcommand{\semanticusl}{\textit{SemanticUSL}}

\newcommand{\audi}{\textit{A2D2}}
\newcommand{\waymo}{\textit{Waymo~Open}}
\newcommand{\libre}{\textit{LIBRE}}
\newcommand{\carla}{\textit{CARLA}}
\newcommand{\gta}{\textit{GTA-V~\acs{lidar}}}
\newcommand{\dense}{\textit{DENSE}}
\newcommand{\pandaset}{\textit{PandaSet}}

\newcommand{\sensortosensor}{\textit{sensor-to-sensor}}
\newcommand{\datasettodataset}{\textit{dataset-to-dataset}}
\newcommand{\daytonight}{\textit{day-to-night}}
\newcommand{\countrytocountry}{\textit{geography-to-geography}}
\newcommand{\weathertoweather}{\textit{weather-to-weather}}
\newcommand{\simtoreal}{\textit{sim-to-real}}
\newcommand{\neartofar}{\textit{near-to-far}}  

\newcommand{\stos}{\textit{S2S}}
\newcommand{\dtod}{\textit{D2D}}
\newcommand{\dton}{\textit{D2N}}
\newcommand{\ctoc}{\textit{G2G}}
\newcommand{\wtow}{\textit{W2W}}
\newcommand{\stor}{\textit{S2R}}

\usepackage{soul}	
\usepackage[normalem]{ulem} 
\usepackage{etoolbox}  
\newtoggle{draft_mode}
\toggletrue{draft_mode}

\togglefalse{draft_mode}

\iftoggle{draft_mode}
{
    \usepackage{todonotes}
    \paperwidth=\dimexpr \paperwidth + 6cm\relax
    \oddsidemargin=\dimexpr\oddsidemargin + 3cm\relax
    \evensidemargin=\dimexpr\evensidemargin + 3cm\relax
    \marginparwidth=\dimexpr \marginparwidth + 3cm\relax
}{
    \usepackage[disable]{todonotes}
}

\newcommand{\todoMD}[2][]{\protect\todo[color=blue!40,#1]{{#2}}}

\makeatletter
		\if@todonotes@disabled

		\else

		\fi
\makeatother

\iftoggle{draft_mode}
{

}{

}

\newcommand{\papertitle}{A Survey on Deep Domain Adaptation for LiDAR Perception}

\title{\LARGE \bf \papertitle}
\author{
Larissa T. Triess$^{1,2,*}$
\and
Mariella Dreissig$^{1}$
\and
Christoph B. Rist$^{1}$
\and
J. Marius Z\"ollner$^{1,3}$%
\thanks{$^{1}$Mercedes-Benz AG, Research and Development, Stuttgart, Germany}%
\thanks{$^{2}$Karlsruhe Institute of Technology (KIT), Karlsruhe, Germany}%
\thanks{$^{3}$Research Center for Information Technology (FZI), Karlsruhe, Germany}%
\thanks{$^{*}$Contact: \texttt{larissa.triess@daimler.com} (\href{https://orcid.org/0000-0003-0037-8460}{0000-0003-0037-8460})}
}

\begin{document}

\maketitle
\thispagestyle{fancy}
\pagestyle{empty}


\begin{abstract}

Scalable systems for automated driving have to reliably cope with an open-world setting.
This means, the perception systems are exposed to drastic domain shifts, like changes in weather conditions, time-dependent aspects, or geographic regions.
Covering all domains with annotated data is impossible because of the endless variations of domains and the time-consuming and expensive annotation process.
Furthermore, fast development cycles of the system additionally introduce hardware changes, such as sensor types and vehicle setups, and the required knowledge transfer from simulation.

To enable scalable automated driving, it is therefore crucial to address these domain shifts in a robust and efficient manner.
Over the last years, a vast amount of different domain adaptation techniques evolved.
There already exists a number of survey papers for domain adaptation on camera images, however, a survey for \acs{lidar} perception is absent.
Nevertheless, \acs{lidar} is a vital sensor for automated driving that provides detailed 3D scans of the vehicle's surroundings.
To stimulate future research, this paper presents a comprehensive review of recent progress in domain adaptation methods and formulates interesting research questions specifically targeted towards \acs{lidar} perception.

\end{abstract}

\FloatBarrier  


\section{Introduction}
\label{sec:introduction}


\begin{figure}[t]
	\centering
	\includegraphics[width=0.96\linewidth]{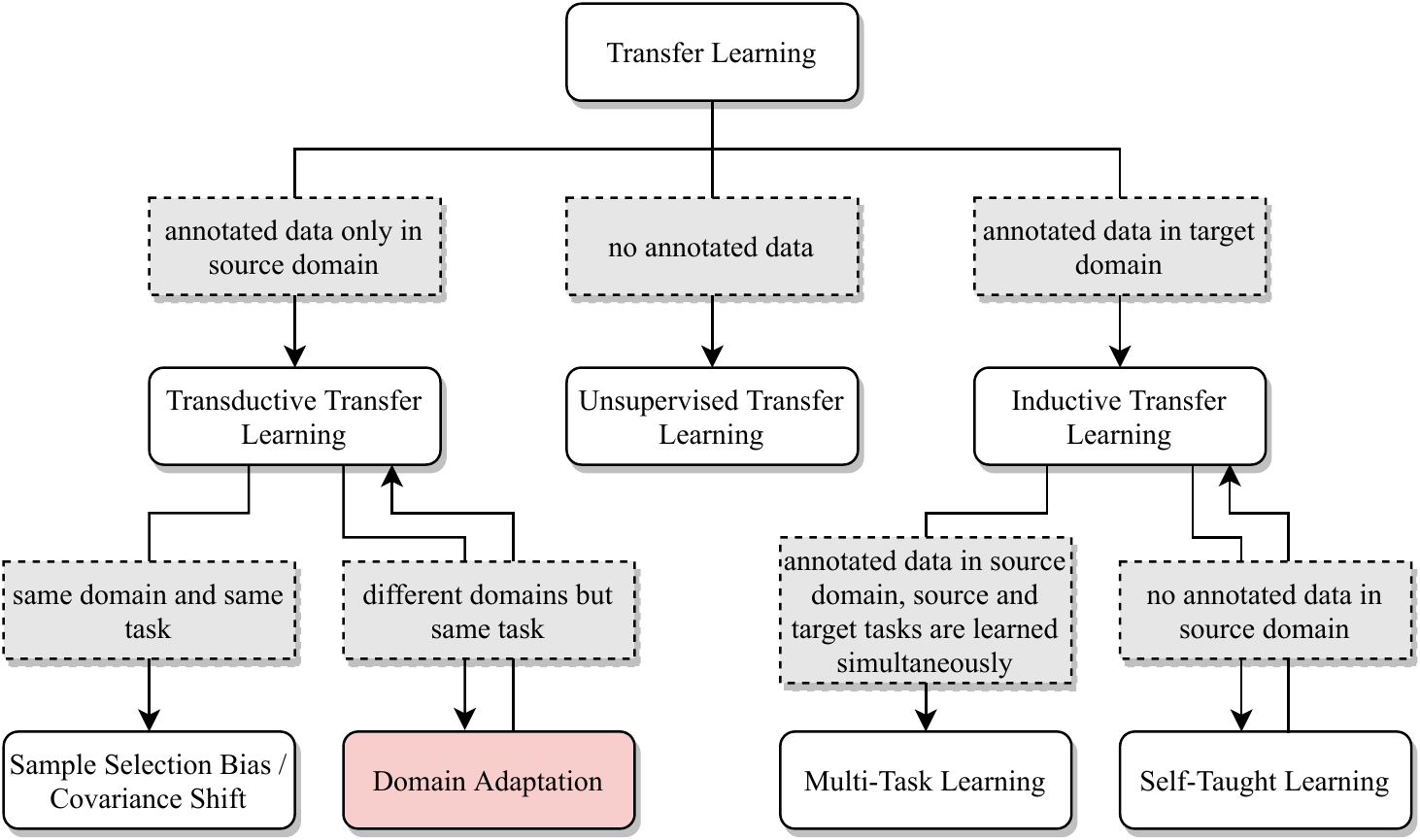}
	\caption{
		\textbf{Overview of Transfer Learning}:
		Domain adaptation is a type of transductive transfer learning where the same task is performed in different, but related domains with annotated data only in the source domain.
		(Figure adapted from~\cite{Pan2010KDE}).
	}
	\label{fig:transfer_learning}
\end{figure}

Highly automated vehicles and robots require a detailed understanding of their dynamically evolving surroundings.
Over the past few years, deep learning techniques have shown impressive results in many perception applications.
They typically require a huge amount of annotated data matching the considered scenario to obtain reliable performances.
A major assumption in these algorithms is that the training and application data share the same feature space and distribution.
However, in many real-world applications, such as in the field of automated driving, this assumption does not hold, since the agents live in an open-world setting.
Furthermore, collection and annotation of large datasets for every new task and domain is extremely expensive, time-consuming, and not practical for scalable systems.
A domain is defined as the scope of application for the algorithm.

A common scenario includes solving a detection task in one domain with training data stemming from another domain.
In this case, the data may differ in their feature space or follow a different data distribution.
Examples for these divergent domains can be different geographical regions, weather scenarios, seasons, other timely aspects, and many more.
For scalable automated driving, fast development cycles, simulated data, and changing sensor setups also play a major role.
Usually, when a perception system is exposed to such a domain shift, its performance drops drastically.
However, it is possible to pass knowledge from a different but related source domain to the desired target domain with transfer learning.
Specifically, \ac{DA} requires no manual annotations to adapt to new domains and therefore promises a cheap and fast solution to deal with domain shifts (compare \figref{fig:transfer_learning}).

In recent years, the research community proposed a vast amount of techniques to transfer knowledge between domains to mitigate the effect of performance drops.
The majority of the proposed methods are targeted towards \ac{DA}~techniques on 2D camera images~\cite{Wilson2020ArXiv,Toldo2020ArXiv,Zhao2020TNNLS}.
Most of these approaches aim for a global feature alignment and ignore local geometric information, which are crucial in 3D perception~\cite{Qin2019NIPS-PointDAN}.
Therefore, recent papers address \ac{DA} specifically for \acs{lidar} perception~\cite{Saleh2019ICCVWorkshop,Rist2019IV,Alonso2020ArXiv,Wu2019ICRA-SqueezeSegV2}.

The focus of this paper is to give an overview on \ac{DA}~methods that specifically address deep learning based \acs{lidar} perception and discuss their unique features, use-cases, and challenges.
The paper is organized as follows:
\secref{sec:background}~gives an introduction to common \acs{lidar} perception tasks and the terminology of~\ac{DA}.
The section also includes an overview on typical baselines, datasets, \ac{DA}~applications, and metrics.
\secref{sec:methods}~categorizes common \ac{DA}~approaches for \acs{lidar}.
In~\secref{sec:discussion}, we discuss different aspects of the presented approaches and give an outlook on interesting research directions.


\section{Background}
\label{sec:background}

Wilson~\etal~\cite{Wilson2020ArXiv} provide an extensive survey on image-based \ac{DA} approaches.
This paper uses similar terminology but extends the \ac{DA} methods with \acs{lidar}-specific categories and focus on \acs{lidar}-related literature.
The following gives an introduction to the building components for deep learning-based \acs{lidar}~\ac{DA} research.

\subsection{\acs{lidar} Perception}
\label{sec:background_lidar-perception}

\acs{lidar} sensors are used in autonomous vehicles to obtain precise distance measurements of the 3D surrounding and extract high-level information about the underlying scenery.
Typical tasks are
object detection, including tracking and scene flow estimation~\cite{Zhou2018CVPR-VoxelNet,Lang2019CVPR-PointPillars,Liu2019CVPR};
point cloud segmentation with semantic segmentation, instance segmentation, and part segmentation~\cite{Milioto2019IROS-RangeNet++,Xu2020ArXiv-SqueezeSegV3,Sirohi2021ArXiv};
and scene completion~\cite{Rist2020IV,Agia2020CoRL}.
Depending on the task at hand and other factors, a number of different processing techniques and data representations for the raw data emerged~\cite{Guo2020PAMI}.
PointNet~\cite{Charles2017CVPR-PointNet} can process generic unordered sets of points and computes features directly from the raw 3D point cloud.
Newer methods either focus on computing features on local regions or exploit grid-based structures to abstract hierarchical relations within the point cloud.
Examples are graph-based~\cite{Yue2019TG-DGCNN,Klokov2017ICCV}, multi-view-based~\cite{Kalogerakis2017CVPR-ShapePFCN}, voxel-based~\cite{Zhou2018CVPR-VoxelNet,Lang2019CVPR-PointPillars}, or higher dimensional lattices~\cite{Su2018CVPR-SPLATNet,Rao2019CVPR-SFCNN}.
For an extensive overview on deep learning methods and representations for 3D point clouds, we refer to~\cite{Guo2020PAMI,Bello2020RS}.

\subsection{Domain Adaptation}
\label{sec:background_domain-adaptation}

\Acf{DA} is a special type of transfer learning~\cite{Pan2010KDE}.
\figref{fig:transfer_learning}~shows the localization of \ac{DA} research in the field of transfer learning.
It is divided into three major categories:
unsupervised transfer learning, where no annotated data is used;
transductive transfer learning, where annotated data is only available in the source domain;
inductive transfer learning, where annotations are available in the target domain.

\Ac{DA} is a type of transductive transfer learning where annotated source data but no annotated target data is available.
It is therefore also called unsupervised domain adaptation in works that use a different terminology~\cite{Wilson2020ArXiv}.
The learning process is defined by performing the same task in different but related domains.
Related domains refer to domains that are placed in a similar setting, such as outdoor driving scenarios, whereas different domains refer to a specific aspect that differs, for example sunny versus rainy days.


\begin{figure}
	\centering

	\includegraphics[scale=1]{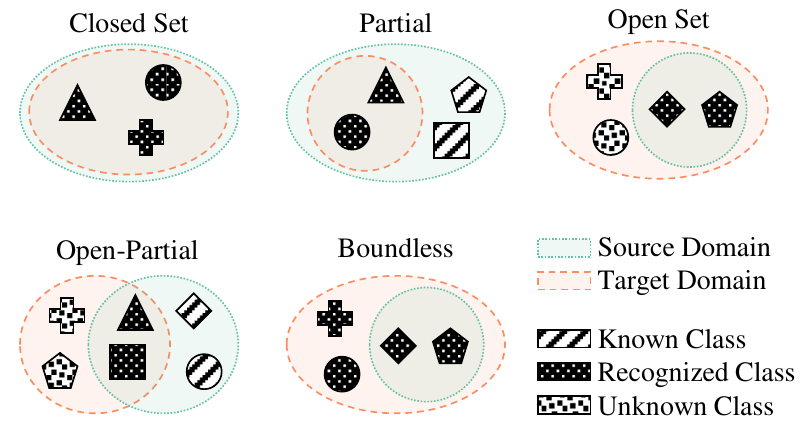}

	\caption{
		\textbf{Intersections between Source and Target Domains}:
		Domain adaptation can be subdivided based on the classes considered in the learning process.
		All classes of the source domain are known and are provided with labels.
		The classes that also occur in the target domain get recognized by the perception network.
		The ones that are not in source, but in target do not get recognized (except for boundless \acs{DA}) and remain unknown.
		(Figure adapted from~\cite{Toldo2020ArXiv}).
	}

	\label{fig:label_set_domains}
\end{figure}

In multi-class classification, \ac{DA} can be subdivided based on the classes of the source and target domains, and on the classes considered in the learning process (\figref{fig:label_set_domains}).
Most papers deal with Closed Set \ac{DA}, where all classes appear in both the source and target domains.
In Partial \ac{DA}, just a subset of the classes from the source domain appears in the target domain.
For Open Set \ac{DA} it is the other way around.
If both sets have both common and unique classes, it is called Open-Partial \ac{DA}.
Boundless \ac{DA} is an Open Set \ac{DA} where all target classes are learned individually.

Wilson~\etal~\cite{Wilson2020ArXiv} suggest a categorization of \ac{DA} methods that reflects the different lines of research in that field.
These include: domain-invariant feature learning, domain-mapping, normalization statistics, ensemble methods, and target discriminative methods.
\secref{sec:methods}~structures the \acs{lidar}-based \ac{DA} approaches into these categories.
To the best of our knowledge, the literature does not provide any works on ensemble methods or target discriminative methods for \acs{lidar}.
Yet, there exist approaches that are specific for \acs{lidar} applications which use domain-invariant data representations.
Therefore, we introduce this additional category.

\subsection{Baselines}
\label{sec:background_baselines}

Compared to \ac{DA} in the camera world, the field of \acs{lidar} \ac{DA} is rather small at this time.
Therefore, many \acs{lidar} papers compare their \ac{DA} approaches to image baselines to compensate the lack of \acs{lidar} baselines.
This section gives a short overview on the baselines used in the presented papers.

The entropy minimization technique is one of the most often referenced baselines.
AdvEnt~\cite{Vu2019CVPR} introduces an adversarial entropy minimization which minimizes the distribution between the source and target based on self-information.
Subsequent work~\cite{Chen2019ICCV} claims that the gradient of the entropy is biased towards samples that are easy to transfer in the entropy minimization approach.
Therefore, they propose a maximum squares loss to balance the gradient of well-classified target samples and prevent the training to be dominated by easy-to-transfer samples.
Minimal-Entropy Correlation Alignment~\cite{Morerio2018ICLR} shows that entropy minimization is induced by the optimal alignment of second order statistics between source and target domains.
On this basis, they propose to use Geodesic instead of Euclidean distances, which improves alignment along non-zero curvature manifolds.

Other image methods that are used as baselines for \ac{DA} are:
CyCADA~\cite{Hoffman2019ICML}, an advanced CycleGAN~\cite{Zhu2017ICCV-CycleGAN};
FeaDA~\cite{Chen2017ICCV}, a joint global and class-specific domain adversarial learning framework;
and OutDA~\cite{Tsai2018CVPR}, a multi-level adversarial network that performs output space \ac{DA} at different feature levels.

\subsection{Datasets}
\label{sec:background_datasets}

Over the last years, numerous \acs{lidar} datasets with various annotation types for automated driving emerged.
A number of these datasets are also used for \ac{DA}.
Most of the works are based on \kitti{}~\cite{Geiger2012CVPR}, \semantickitti{}~\cite{Behley2019ICCV}, \nuscenes{}~\cite{Caesar2020CVPR}, \semanticposs{}~\cite{Pan2020IV}, and \audi{}~\cite{Geyer2020A2D2}.
However, none of these datasets is specifically designed to foster \ac{DA} research and therefore methods using these datasets are hard to compare against each other, due to varying interpretations of the label mappings and the \ac{DA} applications.

Therefore, the following datasets were recently released to address the growing demand in \ac{DA} research.
\semanticusl{}~\cite{Jiang2020ArXiv} is a dataset designed for \ac{DA} between \semantickitti{} and \semanticposs{}.
\libre{}~\cite{Carballo2020IV} and \dense{}~\cite{Bijelic2020CVPR} both provide controlled scenes in adverse weather with multiple sensors and modalities.
The \waymo{} dataset~\cite{Sun2020ArXiv} even includes a \ac{DA} benchmark.

All mentioned datasets are real-world recordings in urban scenarios.
However, synthetic datasets also play a major role in the field of \ac{DA}.
The two most commonly used simulation frameworks are \carla{}~\cite{Dosovitskiy2017CARLA} and \gta{}~\cite{Yue2018ICMR} which enable the simulation of different sensor models.

\subsection{Applications and Use-Cases}
\label{sec:background_usecases}

Using simulators for autonomous driving applications gained a lot of interest in the past years and also increased the research on \simtoreal{} \ac{DA}.
Similarly, \countrytocountry{} \ac{DA} has to address changes in geographical and environmental regions that largely differ in shapes of otherwise similar objects, \eg traffic signs.
Adverse weather conditions, such as fog or rain can substantially deteriorate the detection capabilities of a \acs{lidar}, since laser beams are being reflected and scattered by the droplets or particles in the atmosphere.
Therefore, \weathertoweather{} \ac{DA} considers different weather scenarios and seasons.
In contrast to \acs{DA} for cameras, \daytonight{} \acs{DA} is not important to investigate for \acs{lidar}, since \acs{lidar} is an active sensor that is less dependent on external illumination compared to cameras.

Another important application in the field of development cycles and vehicle setup is the case of \sensortosensor{} \ac{DA}.
It tackles the differences in resolution, mounting position or other sensor characteristics like range of vision, noise characteristics and reflectivity estimates.
Most of the related work considers a far more general case for their research, namely \datasettodataset{}.
It involves multiple of the above mentioned \ac{DA} applications at once.
Several publicly available driving datasets are used to develop and evaluate the adaptation capabilities.
Here, \countrytocountry{} and \sensortosensor{} usually occur at once, often paired with seasonal changes, making this task especially challenging.


\section{Methods}
\label{sec:methods}

This section presents the state of the art on \ac{DA} for \acs{lidar}-based environment perception (\tabref{tab:sota_overview}).
The approaches are either data-driven, such as domain-invariant data representation, domain mapping, and normalization statistics, or model-driven, such as domain-invariant feature learning.


\begin{table}
  \caption{\textbf{Domain Adaptation Methods}}
  \label{tab:sota_overview}
  \setlength{\tabcolsep}{4.7pt}

  {\footnotesize
  $^{a}$Applications:
  \ctoc{}~--~\countrytocountry{},
  \dtod{}~--~\datasettodataset{},
  \dton{}~--~\daytonight{},
  \stos{}~--~\sensortosensor{},
  \stor{}~--~\simtoreal{},
  \wtow{}~--~\weathertoweather{}
  \newline
  $^{b}$Perception:
  detect~--~object detection,
  semseg~--~semantic segmentation
  }

  \begin{tabularx}{\linewidth}{@{} p{2cm} x{1.35cm} x{1.1cm} p{1.7cm} l @{}}
    \toprule
    \textbf{Paper} &
    \textbf{Appl.}$^{a}$ &
    \textbf{Task}$^{b}$ &
    \textbf{Datasets} &
    \textbf{Baselines} \\

    \midrule
    \multicolumn{5}{c}{Domain-Invariant Data Representation (\secref{sec:methods_representation})} \\
    \midrule

    \cite{Triess2019IV}, \cite{Shan2020ArXiv}, \cite{Elhadidy2020NILES} &
    \stos{} &
    semseg$^{\dagger\dagger}$ &
    \cite{Geiger2012CVPR}, \cite{Behley2019ICCV}, \cite{Dosovitskiy2017CARLA} &
    - \\

    PiLaNet~\cite{Piewak2019ITSC} &
    \stos{} &
    semseg &
    custom &
    - \\

    Complete~\!\&\!~Label~\cite{Yi2020ArXiv} &
    \dtod{} &
    semseg &
    \cite{Geiger2012CVPR}, \cite{Caesar2020CVPR}, \cite{Sun2020ArXiv} &
    \cite{Chen2017ICCV}, \cite{Tsai2018CVPR} \\

    \midrule
    \multicolumn{5}{c}{Domain Mapping (\secref{sec:methods_domain-mapping})} \\
    \midrule

    \cite{Saleh2019ICCVWorkshop}, \cite{Sallab2019ICMLWORK}, \cite{Sallab2019NIPSWORK} &
    \dtod{}, \stor{} &
    detect &
    \cite{Geiger2012CVPR}, \cite{Dosovitskiy2017CARLA} &
    - \\

    \cite{Caccia2019IROS}, \cite{Nakashima2021ArXiv} &
    \stor{} &
    - &
    \cite{Geiger2012CVPR}, \cite{Mozos2019IJRR} &
    - \\

    ePointDA~\cite{Zhao2021AAAI} &
    \stor{} &
    semseg &
    \cite{Geiger2012CVPR}, \cite{Behley2019ICCV}, \cite{Yue2018ICMR} &
    - \\

    Alonso~\etal~\cite{Alonso2020ArXiv}$^\dagger$ &
    \dtod{} &
    semseg &
    \cite{Behley2019ICCV}, \cite{Pan2020IV}, \cite{Roynard2017IJRR} &
    \cite{Vu2019CVPR}, \cite{Chen2019ICCV} \\

    Langer~\etal~\cite{Langer2020IROS} &
    \stos{} &
    semseg &
    \cite{Behley2019ICCV}, \cite{Caesar2020CVPR} &
    (\cite{Morerio2018ICLR}) \\

    \midrule
    \multicolumn{5}{c}{Domain-Invariant Feature Learning (\secref{sec:methods_domain-invariant-approaches})} \\
    \midrule

    SqueezeSeg-V2~\cite{Wu2019ICRA-SqueezeSegV2} &
    \stor{} &
    semseg &
    \cite{Geiger2012CVPR}, \cite{Yue2018ICMR} &
    - \\

    xMUDA~\cite{Jaritz2020CVPR} &
    \ctoc{}, \dtod{}, \dton{} &
    semseg &
    \cite{Behley2019ICCV}, \cite{Caesar2020CVPR}, \cite{Geyer2020A2D2} &
    \cite{Vu2019CVPR}, \cite{Morerio2018ICLR} \\

    LiDARNet~\cite{Jiang2020ArXiv} &
    \dtod{} &
    semseg &
    \cite{Behley2019ICCV}, \cite{Pan2020IV}, \cite{Jiang2020ArXiv} &
    \cite{Hoffman2019ICML} \\

    Wang~\etal~\cite{Wang2019ICCVWORK} &
    \stos{} &
    detect &
    \cite{Geiger2012CVPR}, \cite{Caesar2020CVPR} &
    - \\

    SF-UDA3D~\cite{Saltori20203DV} &
    \dtod{} &
    detect &
    \cite{Geiger2012CVPR}, \cite{Caesar2020CVPR} &
    \cite{LI2017ICLRWORKSHOP} \\

    \midrule
    \multicolumn{5}{c}{Normalization Statistics \& Other Methods (Sections \ref{sec:methods_normalization-statistics} \& \ref{sec:methods_other-methods})} \\
    \midrule

    Rist~\etal~\cite{Rist2019IV}$^\dagger$ &
    \stos{} &
    semseg, detect &
    \cite{Geiger2012CVPR}, custom &
    - \\

    Caine~\etal~\cite{Caine2021ARXIV} &
    \ctoc{}, \wtow{} &
    detect &
    \cite{Sun2020ArXiv} &
    - \\

    \bottomrule
  \end{tabularx}

  {\footnotesize
  $^\dagger$these approaches use a combination with methods from \secref{sec:methods_representation}\newline
  $^{\dagger\dagger}$primary task is upsampling, but \cite{Triess2019IV} and \cite{Elhadidy2020NILES} also test for semseg
  }

\end{table}

\subsection{Domain-Invariant Data Representation}
\label{sec:methods_representation}

\begin{figure}
	\centering

  \begin{subfigure}{\linewidth}
		\centering
		\includegraphics[scale=1]{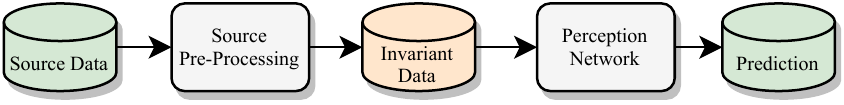}
		\caption{\label{fig:domain_invariant_data_train}Training}
	\end{subfigure}

  \vspace{1em}

	\begin{subfigure}{\linewidth}
		\centering
		\includegraphics[scale=1]{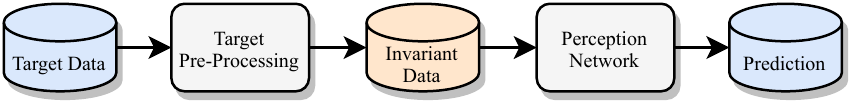}
		\caption{\label{fig:domain_invariant_data_test}Testing}
	\end{subfigure}

	\caption{
		\textbf{Domain-Invariant Data Representation}:
		The data from the source domain at train-time~(\subref{fig:domain_invariant_data_train}) and the data from the target domain at test-time~(\subref{fig:domain_invariant_data_test}) are both converted into a hand-crafted common representation prior to being fed to the perception pipeline.
	}

	\label{fig:domain_invariant_data}
\end{figure}

A domain-invariant representation is a hand-crafted approach to move different domains into a common representation.
\figref{fig:domain_invariant_data}~shows that this approach is basically a data pre-processing after which a regular perception pipeline starts.
It is mostly used to account for the \sensortosensor{} domain shift and receives special attention in \acs{lidar} research.
Available sensors vary in their resolution and sampling patterns while resulting point clouds are additionally influenced by the mounting position and the recording rate of the sensor.
Consequently, the acquired data vary considerably in their statistics and distributions.

This data distribution mismatch makes it unfeasible to apply the same model to different sensors in a naive way.
Therefore, many simple \ac{DA} methods either align the sampling differences in 2D space or use representations in 3D space that are less prone to domain differences.
Other methods include a normalization of the input feature spaces with respect to different mounting positions by spatial augmentations and replacing absolute \acs{lidar} coordinates with relative encoding schemes~\cite{Rist2019IV, Alonso2020ArXiv}.


\subsubsection{Sampling Alignment in 2D Space}
\label{sec:methods_representation-sampling}

The sensor view of a rotating \acs{lidar} scanner resembles a 2D image.
The vertical resolution equals the number of layers and the horizontal resolution depends on the revolution frequency of the sensor.
Convolutional neural networks are often used for \acs{lidar} perception in 2D space.
Even though fully convolutional networks operate independently of the input size, the receptive field still changes with varying sensor resolution.
A straightforward way to align these characteristics manually is by either up-sampling the data~\cite{Triess2019IV,Shan2020ArXiv,Elhadidy2020NILES} or by dropping scan lines~\cite{Alonso2020ArXiv}.

\subsubsection{Geometric Representation in 3D Space}
\label{sec:methods_representation-volumetric}

Geometric representations exploit the inherent 3D structure of point clouds.
In the literature, there are two approaches on that: employing a volumetric (column-like or voxel) data representation or recovering the geometric 3D surface.

PointPillars~\cite{Lang2019CVPR-PointPillars} represents \acs{lidar} data in a column-like structure.
The authors of~\cite{Piewak2019ITSC} analyze this representation with respect to its \sensortosensor{} \ac{DA} capabilities and conclude that this representation can be applied regardless of the sensor's vertical resolution, as it reduces that axis to one.

Complete~\!\&\!~Label~\cite{Yi2020ArXiv} introduces a more sophisticated approach by exploiting the underlying scenery of the scan which is independent of the recording mechanism.
The method includes a sparse voxel completion network to recover the underlying surface that was sampled by a \acs{lidar}.
This high-resolution representation is then used as an input to a labeling network which provides semantic predictions independent of the domain where the scene originates from.

\subsection{Domain Mapping}
\label{sec:methods_domain-mapping}


\begin{figure}
	\centering

  \begin{subfigure}{\linewidth}
		\centering
		\includegraphics[scale=1]{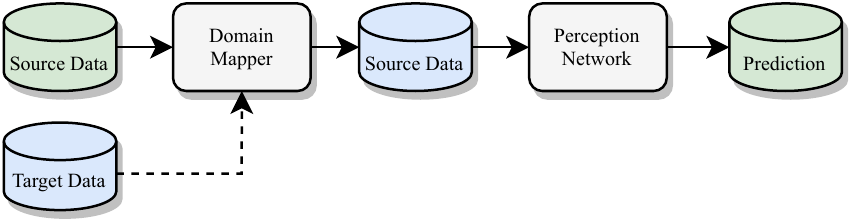}
		\caption{\label{fig:domain_mapping_train}Training}
	\end{subfigure}

  \vspace{1em}

	\begin{subfigure}{\linewidth}
		\centering
		\includegraphics[scale=1]{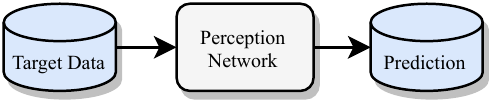}
		\caption{\label{fig:domain_mapping_test}Testing}
	\end{subfigure}

	\caption{
		\textbf{Domain Mapping}:
		This is the most commonly used configuration for domain mapping.
		During training~(\subref{fig:domain_mapping_train}) the labeled source data is conditionally (dashed line) mapped to the target domain where a perception network is trained.
		At test time~(\subref{fig:domain_mapping_test}), the trained perception network can directly be applied to the target data.
	}

	\label{fig:domain_mapping}
\end{figure}

Domain mapping aims at transferring the data of one domain to another domain and is most often used in \simtoreal{} and \datasettodataset{} applications.
\figref{fig:domain_mapping}~shows a typical setup for domain mapping.
Annotated source data is usually transformed to appear like target data, creating a labeled pseudo-target dataset.
With the transformed data, a perception network is trained which can then be applied to target data at test time.

For images, domain mapping is usually done adversarially and at pixel-level in the form of image-to-image translation with conditional GANs~\cite{Choi2018CVPR,Royer2020XGAN,Benaim2017NIPS,Taigman2017ICLR}.
Similar principles apply to \acs{lidar} data, however, there also exists a number of methods that do not rely on adversarial training.

\subsubsection{Adversarial Domain Mapping}
\label{sec:methods_domain-mapping_adversarial}

Adversarial domain mapping is typically accomplished with conditional GANs~\cite{Mirza2014CGAN}.
The generator translates a source input to the target distribution without changing the underlying semantic meaning.
A perception network can then be trained on the translated data using the known source labels.
The translated data shall have the same appearance as the target data.

In contrast to methods for camera images or generic point clouds, the number of papers that adversarially generate realistic \acs{lidar} data is very limited.
In fact, most approaches use unmodified image GANs, such as CycleGAN~\cite{Zhu2017ICCV-CycleGAN}, and apply them to top-view projected images of the \acs{lidar} scans~\cite{Saleh2019ICCVWorkshop,Sallab2019ICMLWORK,Sallab2019NIPSWORK}.
In all three of these works, the images are translated between synthetic and real-world domains.
They evaluate their \ac{DA} capabilities on top-view object detection for which they use YOLOv3~\cite{Redmon2018ArXiv-YOLOv3}.
They show an improvement for object detection when YOLOv3 is trained with the domain adapted data.

Another possibility is to exploit the sensor-view image of the \acs{lidar}.
In contrast to the top-view projection, the sensor-view image is a dense and lossless representation with which the original 3D point cloud can be recovered.
Caccia~\etal~\cite{Caccia2019IROS} provide an unsupervised method for both conditional and unconditional \acs{lidar} generation in projection space and test their method on reconstruction of noisy data.
The generated data does not possess any point-drops as they usually occur in real-world data.
Therefore, DUSty~\cite{Nakashima2021ArXiv} additionally incorporates a differentiable framework that can sample binary noises to simulate these point-drops and mitigate the domain gap between the real and synthesized data.
Similarly, ePointDA~\cite{Zhao2021AAAI} learns a dropout noise rendering from real data and applies it to synthetic data.

\subsubsection{Non-Adversarial Domain Mapping}
\label{sec:methods_domain-mapping_traditional}

The non-adversarial mapping techniques primarily focus on the sampling and distribution differences between \acs{lidar} sensors.

Alonso~\etal~\cite{Alonso2020ArXiv} address the \datasettodataset{} problem by using a data and class distribution alignment strategy.
In the data alignment process, a number of simple augmentation techniques, such as $xyz$-shifts, replacing absolute with relative features, and the dropping of \acs{lidar} beams, are used.
In the class distribution alignment, it is assumed that the target domain has a similar class distribution as the source domain, since the datasets in both domains are recorded in urban scenarios.
Therefore, the Kullback-Leibler divergence between the two class distributions is minimized.

Another approach is to use a re-sampling technique to address \sensortosensor{} domain shifts~\cite{Langer2020IROS}.
Here, several scans of a recorded sequence are accumulated over time.
The specifications of a \acs{lidar} sensor with lower resolution are then used to sample a single semi-synthetic scan from the accumulated scene.
In the second step of the domain transfer, the semantic segmentation model has to be re-trained with geodesic correlation alignment to align second-order statistics between source and target domains to generalize to a \datasettodataset{} setting~\cite{Morerio2018ICLR,Wu2019ICRA-SqueezeSegV2}.
To quantitatively verify the effectiveness of their method, the authors labeled the target dataset with the same classes as their source dataset.

\subsection{Domain-Invariant Feature Learning}
\label{sec:methods_domain-invariant-approaches}

State-of-the-art methods in domain-invariant feature learning employ a training procedure that encourages the model to learn a feature representation that is independent of the domain.
This is done by finding or constructing a common representation space for the source and target domain.
In contrast to domain-invariant data representations, these approaches are not hand-crafted but use learned features.

If the classifier model performs well on the source domain using a domain-invariant feature representation, then the classifier may generalize well to the target domain.
The basic principle is depicted in~\figref{fig:domain_invariant_features}.
Common approaches for domain-invariant feature learning can be categorized into two basic principles:


\begin{figure}
	\centering

  \begin{subfigure}{\linewidth}
		\centering
		\includegraphics[scale=1]{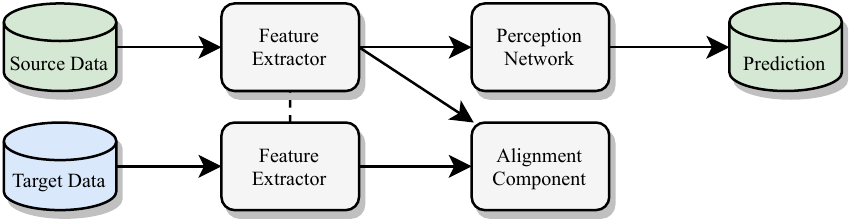}
		\caption{\label{fig:domain_invariant_features_train}Training}
	\end{subfigure}

  \vspace{1em}

	\begin{subfigure}{\linewidth}
		\centering
		\includegraphics[scale=1]{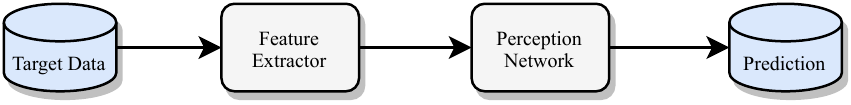}
		\caption{\label{fig:domain_invariant_features_test}Testing}
	\end{subfigure}

	\caption{
		\textbf{Domain-Invariant Feature Learning}:
		A feature extractor network and an alignment component learn a domain-invariant feature encoding~(\subref{fig:domain_invariant_features_train}).
		At test time~(\subref{fig:domain_invariant_features_test}), the domain-invariant feature extractor is applied to the target data.
	}

	\label{fig:domain_invariant_features}
\end{figure}

\subsubsection{Divergence Minimization}
\label{sec:divergence_minimization}

One approach on creating a domain-invariant feature encoder is minimizing a suitable divergence measure between the representation of the source and the target domain within the \acl{DNN}.
SqeezeSegV2~\cite{Wu2019ICRA-SqueezeSegV2} propose a \ac{DA} pipeline where the discrepancies in the batch statistics from both domains are minimized.
Adapting the work of Morerio~\etal~\cite{Morerio2018ICLR}, they use a loss function based on the geodesic distance between the output distributions.
The authors apply their presented approach to a \simtoreal{} setting.
xMUDA~\cite{Jaritz2020CVPR} minimizes the domain gap by incorporating a learning structure which utilizes perception algorithms on both 2D images and 3D point clouds.
A specific cross-modal loss is designed as a divergence measure.
Thus, an information exchange between the two modalities benefits the overall performance in presence of a domain shift.
The method is tested on \daytonight{}, \countrytocountry{} and \datasettodataset{} scenarios.

\subsubsection{Discriminator-based Approaches}
\label{sec:discriminator}

The basic idea of domain-invariant feature learning with a discriminator is to utilize adversarial training to force the feature encoder to learn only domain-invariant features.
The feature extractor learned a domain-invariant feature representation as soon as the discriminator is not able to distinguish from which domain the feature representation originated (i.e. source or target).
LiDARNet~\cite{Jiang2020ArXiv} follows this discriminator approach, which was previously applied to camera images in~\cite{Zhao2019PMLR}.
The same principle of minimizing domain gaps by employing a discriminator is applied in~\cite{Wang2019ICCVWORK}.
Here, the authors conduct the model adaptation on intermediate layers of the \acs{DNN} to improve the detection of far range objects.
Though this \neartofar{} is a special case of \ac{DA} within one \acs{lidar} sensor, the effectiveness of this approach is demonstrated in \datasettodataset{} use cases.

The authors of~\cite{Saltori20203DV} present a different but interesting take on \ac{DA} for \acs{lidar} point clouds in their work.
They exploit temporal consistency in the detection to generate pseudo-labels on the target domain.
Following, they built a model which does not rely on annotations from the source domain by utilizing a variant of self-taught learning.

\subsection{Normalization Statistics}
\label{sec:methods_normalization-statistics}

The primary use of normalization techniques is to improve training convergence, speed and performance.
These advantages have initially been verified on image datasets, image related tasks and their respective model architectures.
On \ac{lidar} data, normalization techniques are used equally for improved feature extraction for a variety of tasks~\cite{Wu2019ICRA-SqueezeSegV2,Zhou2018CVPR-VoxelNet,Lang2019CVPR-PointPillars,Charles2017CVPR-PointNet,Su2018CVPR-SPLATNet}.
On images, the properties of normalization are used for explicit \ac{DA}.
A set of normalization statistics per domain are expected to separate domain knowledge from task knowledge.
This encourages \acp{DNN} to learn style-invariant representations of input data.
Conversely, manipulating the distribution of intermediate layer activations is explicitly used for image style transfer.
However, experimental studies that verify a strong similarity between style normalization on camera images and sensor or scene normalization on \acs{lidar} are absent.

In \acp{DNN}, normalization layers improve training convergence by aligning the distributions of training data and therefore controlling internal covariate shift and the scale of the gradients.
Distribution alignment is implemented by a normalization of mean and variance of activations over partitions of the batch-pixel-feature tensor.
The most prominent normalization technique is batch normalization~\cite{Ioffe2015ICML}.
Building on the same basic idea, subsequent related normalization procedures~\cite{Wu2018ECCV,nam2018batchinstance,Ulyanov2016ArXiv} address issues
with batch normalization related to implementation, network architectures, or certain data domains.

Adaptive batch normalization is a simple and straightforward \ac{DA} method that re-estimates the dataset statistics on the target domain as a natural extension of the batch norm approach~\cite{LI2017ICLRWORKSHOP}.
However, this normalization approach alone does not lead to a satisfactory object detection
performance in a \acs{lidar} \sensortosensor{} \ac{DA} setup~\cite{Rist2019IV}.
When training on multiple image data domains simultaneously,
switching between per-domain statistics is used for \ac{DA} on image tasks as the second stage of a two-stage approach~\cite{Chang2019CVPR}.
Initial pseudo-labels are iteratively refined using separate batch norm statistics for each domain.
The effectiveness of per-domain statistics on \acs{lidar} domain gaps has not been verified experimentally.

\subsection{Other Methods}
\label{sec:methods_other-methods}

Recently, it is demonstrated that teacher-student knowledge distillation can increase performance on a target domain in an unsupervised \ac{DA} setting on \acs{lidar}~\cite{Caine2021ARXIV}.
A teacher network is trained on the source domain and creates pseudo-labels on the target domain.
The smaller student network is then trained on source and target domains simultaneously.
A setup to adapt to a domain with different geometries (\datasettodataset{}) is used as experimental evaluation.
The pseudo-label trained student networks show better generalization capabilities on the target domain than the teacher networks.
Practitioners benefit from the simplicity of the approach and the option to adhere to inference time budgets with small student networks.
However, in practice the sensor setup itself might be different instead of only the scene geometry.
This yields a more difficult problem as the networks need to work on both domains simultaneously.


\section{Discussion}
\label{sec:discussion}

After reviewing the recent advances in the related literature, we discuss the main challenges that remain open in \ac{DA} for \acs{lidar} perception.
They pose interesting research directions for future works.

\subsection{Comparability and Transfer from other Modalities}
\label{sec:discussion_comparison}

An essential part of research is the comparability between different approaches to foster further research in promising directions.
In most papers, the success of the \ac{DA} process is measured by the performance of a downstream perception task.
However, this has two major drawbacks.
First, it is assumed that the quality of the domain adaptation process directly correlates with the performance changes in the downstream perception.
However, there is no proof for this yet.
Second, the methods are still not comparable, since they all use different datasets, task settings, label sets, and report different metrics.
One reason might be that most of the advanced \ac{DA} approaches in \acs{lidar} only emerged recently, therefore no metric or baseline prevailed.

To mitigate the lack of a \acs{lidar}-specific baseline, many of the presented works use image-based approaches as a baseline comparison for their own work.
However, these are usually unfair evaluations, since the baseline methods are not optimized for \acs{lidar} data.
Furthermore, it is unknown whether there is a qualitative difference in the domain gaps between image domains (camera models, image styles) and \acs{lidar} domains (\acs{lidar} sensors, 3D scenes).
If so, this would prevent all the \ac{DA} methods for images to be further optimized with respect to \acs{lidar} applications.

\subsection{Discrepancies in Domain Gap Quality}
\label{sec:discussion_image-approaches}

The size of a domain gap can be measured in terms of model performance on a given task if target labels are available.
Consequently, this measure of the apparent size of a domain gap is model-specific and task-specific.
Nevertheless, a change of the \ac{lidar} sensor causes a more severe impact on the final performance consistently across several tasks and models than changes in weather or location.
It seems that a \sensortosensor{} domain gap is not easily covered by learned and implicit \ac{DA} methods such as normalization statistics and adversarial domain mapping that work well in the image domain.
Hand-crafted \ac{DA} methods based on the explicit geometric properties of \acs{lidar} data and their representation are the only ones that yield reasonable results on \sensortosensor{} domain shifts so far.
We see a qualitative difference between \ac{DA} for \acs{lidar} and images within the state of the art.

\subsection{Relevance of Cross-Sensor Adaptation}
\label{sec:discussion_cross-sensor}

Various sensor types were developed in the past decade.
Available \acs{lidar} sensors mainly vary in their design and functionality, i.e. rotating and oscillating \acp{lidar}~\cite{Li2020, Royo2019}.
Consequently, the scan-pattern and thus the data representation differs considerably between the sensors.

Although the mechanical spinning \acs{lidar} is mainly used for perception in the autonomous driving research~\cite{Behley2019ICCV,Caesar2020CVPR}, all sensor types have their benefits.
The publicly available \pandaset{} dataset~\cite{Pandaset} is one of the first datasets to incorporate two different sensor types: a $360^\circ$ covering spinning \acs{lidar} and an oscillating \acs{lidar} with a snake scan-pattern.
Since it is unclear which type of sensor will prevail in the context of autonomous driving in the future, it is crucial to advance the development of sensor-invariant perception systems.
The in \secref{sec:methods_representation} presented geometric and volumetric data representation approaches are valuable to fully benefit from the different sensor types.
Constructing a domain-invariant data representation in an intermediate step makes it possible to re-use already trained \acp{DNN} and apply them to new sensor setups.
This is essential to lower the research costs and to keep up with the sensor development.

\subsection{Adaptation in Different Weather Scenarios}
\label{sec:discussion_weather}

\acs{lidar} sensors are heavily impacted by adverse weather conditions, such as rain or fog, which cause undesired measurements and leads to perception errors.
There exists some work on denoising \acs{lidar} perception~\cite{Heinzler2020RAL}, but the subject is not specifically tackled in a \ac{DA} setting.
With the release of new datasets containing adverse weather scenarios~\cite{Carballo2020IV,Bijelic2020CVPR}, we believe it is possible to foster the research for \weathertoweather{} applications.

\subsection{Generative Models for Domain Translation}
\label{sec:discussion_generative}

The category of adversarial domain mapping techniques (\secref{sec:methods_domain-mapping_adversarial}) includes only four approaches, none of which is capable to generate realistic \acs{lidar} point clouds in 3D (they use top-view projections of the point clouds).
This is surprising, since the same strategy is thriving in the image world, where a lot of research is conducted to generate realistic images.
There are two possible explanations to it: either, it is simply not necessary to use generative models to create realistic point clouds, in the sense that other approaches are far more powerful, or it is not possible to achieve the required high quality of the generated data.
Either way, up until now there exists no study that either proves or contradicts any of these assumptions.
A method similar to~\cite{Hubschneider2020} that analyzes the \acs{DA} performance of generative models for \acs{lidar} domain mapping might help to advance research in this direction.

\subsection{Open-Partial Domain Adaptation}
\label{sec:discussion_open-partial}

The majority of the presented approaches deal with \datasettodataset{} applications, a manifold adaptation task, where both the sensor type and the environment change.
This makes the task particularly difficult.
However, another effect comes to light.
Usually, datasets have unique labeling strategies and include a different set of classes.
For example, \semantickitti{} has 28 semantic classes while \nuscenes{} divides into 32 classes.
Since the domains have both common and unique classes, this is called an Open-Partial \ac{DA} problem~\cite{Toldo2020ArXiv}.

However, none of the presented approaches directly tackles the Open-Partial formulation, but rather perform a label mapping strategy that allows them to address this as a Closed Set \ac{DA} problem.
Some of the segmentation approaches do not focus on segmenting the entire scenery from the start, but only perform foreground segmentation, for example cars versus background~\cite{Yi2020ArXiv}.
A common strategy when semantically segmenting entire scenes, is to find a common minimal class mapping between the two datasets, that discards all classes that cannot be matched~\cite{Alonso2020ArXiv}.
Some works even re-label one of the datasets to perfectly match the label definitions of the other dataset~\cite{Langer2020IROS}.

In the image world, there are already works that deal with the Open-Partial and Open-Set \ac{DA} problems~\cite{Busto2017ICCV,Saito2018ECCV,Luo2020ICML}.
In the \acs{lidar} world, this field still holds a lot of potential for future research.
We believe that a good strategy for these questions can help to advance in scalable systems for automated vehicles.
\todo{Reviewer 3: With regard to the autonomous driving application: What is the technology readiness based on the existing work? What is easy to detect and where are technology gaps that must be filled? Where are the corner cases where LiDAR comes to its border? What are typical difficulties that appear in the papers?}
\todoMD{Do we want to include details about some of these questions?}


\section{Conclusion}
\label{sec:conclusion}

This paper reviewed several current trends of \ac{DA} for \acs{lidar} perception.
The methods are classified into five different settings: domain-invariant feature learning, domain mapping, domain-invariant data representations, normalization statistics and other methods.
Depending on the application and the extent of the domain gap itself, different types of \ac{DA} approaches can be used to bridge the gap.
Especially the hand-crafted \ac{DA} methods like normalization statistics and domain-invariant feature representation can be used to complement \ac{DA} learning methods like domain mapping and domain-invariant feature learning.

The main contribution of this paper is to formulate several interesting research questions.
Addressing them can benefit the \ac{DA} research for \acs{lidar} in the future.
The introduction of a prevailing \acs{lidar} \ac{DA} benchmark can yield an important step forward to make existing and new works comparable.
Furthermore, many of the surveyed approaches stem from ideas developed for cameras.
The adaptation to the \acs{lidar} modality yields promising results, but the characteristics of the \acs{lidar} sensor pose novel modality-specific challenges and research questions.
Examples are the need to address critical use-cases like cross-sensor and weather domain shift.
On the other hand, \daytonight{} adaptation becomes insignificant for \acs{lidar}.
Unlike in the image domain, there exist only few adversarial methods that focus on generating realistic \acs{lidar} point clouds.
Finally, in both camera and \acs{lidar}, research mostly focuses on closed-set \ac{DA}, however for truly scalable automated driving it is important to address the open-partial \ac{DA} problem.

\addtolength{\textheight}{-0.5cm}   

\section*{Acknowledgment}

This work was presented at the \emph{Workshop on Autonomy at Scale} (WS52), IV2021
The research leading to these results is funded by the German Federal Ministry for Economic Affairs and Energy within the project ``KI Delta Learning'' (F\"orderkennzeichen 19A19013A).

{\small
\bibliographystyle{bibliography/IEEEtran}
\bibliography{bibliography/bib_short,bibliography/refs}
}

\end{document}